\documentclass{article}  
\usepackage{subcaption}
\usepackage{amsmath}
\usepackage{amssymb}
\usepackage{float}
\usepackage{url}
\usepackage{xcolor}
\usepackage{graphicx}

\title{\LARGE \bf Extending Monocular Visual Odometry to\\ Stereo Camera Systems by Scale Optimization}

\author{Jiawei Mo$^{1}$ and Junaed Sattar$^{2}$
\thanks{The authors are with the Department of Computer Science and Engineering, University of Minnesota Twin Cities, Minneapolis, MN, USA.
{\tt\small \{$^{1}$moxxx066, $^{2}$junaed\} at umn.edu.}}
}

\begin{document}

\maketitle

\begin{abstract}
This paper proposes a novel approach for extending monocular visual odometry to a stereo camera system. The proposed method uses an additional camera to accurately estimate and optimize the scale of the monocular visual odometry, rather than triangulating 3D points from stereo matching. Specifically, the 3D points generated by the monocular visual odometry are projected onto the other camera of the stereo pair, and the scale is recovered and optimized by directly minimizing the photometric error. It is computationally efficient, adding minimal overhead to the stereo vision system compared to straightforward stereo matching, and is robust to repetitive texture. Additionally, direct scale optimization enables stereo visual odometry to be purely based on the direct method. Extensive evaluation on public datasets (\textit{e.g.}, KITTI), and outdoor environments (both terrestrial and underwater) demonstrates the accuracy and efficiency of a stereo visual odometry approach extended by scale optimization, and its robustness in environments with challenging textures.
\end{abstract}

\section{Introduction}
\label{sec:introduction}
Localization is an essential feature for autonomous robot navigation; however, it can be challenging in certain environments such as indoors and underwater where GPS signals are unavailable or unique landmarks are difficult to detect. Visual odometry (VO) has been widely used for robot localization, which estimates \textit{ego-motion} using only camera(s). Cameras are \textit{passive} sensors and thus consume less energy compared to \textit{active} sensors such as sonar or laser range-finder (\textit{i.e.}, LiDAR). Mobile robots, particularly those operating outdoors or in unstructured domains, benefit greatly from efficient energy usage as it extends the length of deployments, and also reduces downtime between missions.

Depending on the number of cameras in the system, visual odometry can be categorized into \textit{monocular} or \textit{multi-camera} system. Among multi-camera VO, stereo VO is the most widely used. The classic procedure of a stereo VO starts with \textit{stereo matching}. Stereo matching searches feature correspondences between stereo frames; 3D positions of objects are then estimated instantly by triangulation. Subsequently, the camera pose (position and orientation) is estimated with respect to the 3D points. Since the 3D points are fully recovered, so is the camera pose. However, stereo matching can be computationally expensive. For each feature point, its correspondence is found by searching for the most similar patch along the epipolar line exhaustively. Additionally, many stereo matching algorithms will rectify the stereo pair first, which is also very time-consuming. Another challenge is that when the texture is repetitive and of high-frequency, there could be more than one similar patch that would give rise to ambiguity in best-match determination. These scenes are not uncommon outdoors and are often encountered by field robots, such as underwater or mine-exploration robots.

\begin{figure}[t]
    \centering
    \includegraphics[width=\textwidth]{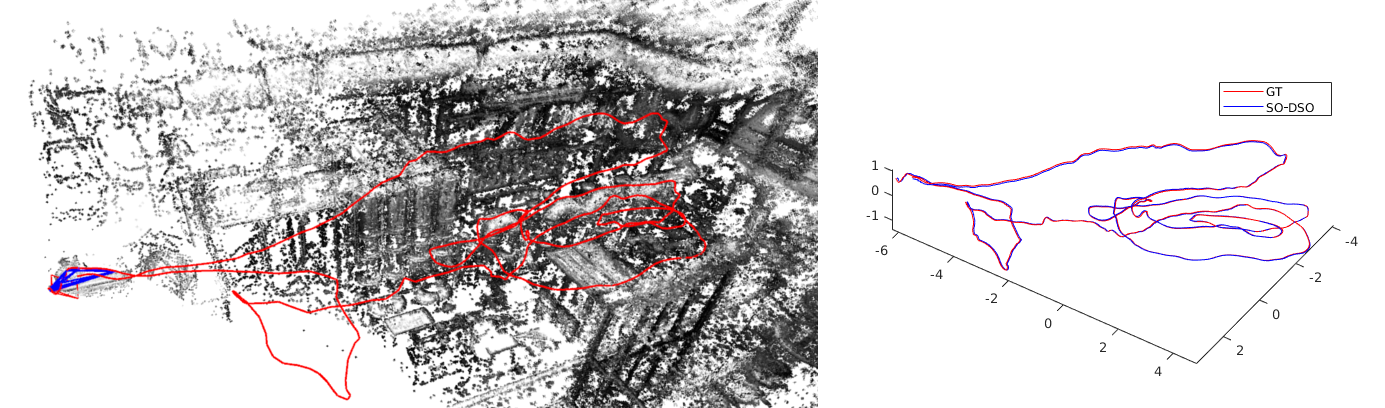}
    \caption{A demonstration of a stereo VO using the proposed method running on MH01 of EuRoC dataset. Left image shows the trajectory and 3D points. Right image compares the trajectory against ground truth.}
    \label{fig:mh01}
\end{figure}

On the other hand, without the need to match points among different cameras, monocular VO algorithms (\textit{e.g.,}~\cite{engel2018direct, forster2014svo, mur2017orb}) are capable of camera tracking in these repetitive scenes and are computationally less expensive than multi-camera VO. As an example, SVO~\cite{forster2014svo} needs only about 3 milliseconds to process each frame. Low-power platforms such as micro air vehicles benefit greatly from computational efficiency. However, monocular VO is not able to fully estimate camera pose. As the camera projects 3D objects onto 2D images, the distance to (\textit{i.e.}, the depth of) the object is lost during this process. For monocular VO, depth is partially recovered from parallax by moving the camera temporally. However, since the camera movement is unknown as well, both depth and camera pose are estimated up to an unknown scale. The detailed discussion about the unknown scale is found in~\ref{sec:mono_vo}. Additionally, the scale tends to drift so that it is inconsistent throughout the process. Scale awareness is important for a number of robotic behaviors including, but not limited to, vision-based control and path planning.

To solve the scale problem without intensive computational cost, authors~\cite{leutenegger2015keyframe,bloesch2015robust,qin2018vins} have fused monocular VO with an inertial measurement unit (IMU) to create visual-inertial navigation systems (VINS). In this case, the IMU provides scale estimation. However, IMU pose propagation is sensitive to measurement noise, thus visual measurements are used to correct the propagated pose. VINS achieve high accuracy and efficiency with a reliable IMU, which needs to be initialized at the beginning. 

In this work, we propose a novel approach to solve the scale problem of monocular VO by incorporating an additional camera rather than an IMU. It combines the strengths of stereo and monocular VO in terms of accuracy and performance. Camera poses and 3D points are estimated by a monocular VO running on one camera; the other camera is only used to address the scale problem by projecting the 3D points from the monocular VO onto it. The optimal scale is solved by minimizing the photometric error in stereo projection. The main contributions of this work are the following: 
\begin{itemize}
  \item A novel algorithm to extend monocular VO to stereo,
  \item Full estimation of camera poses and 3D points with optimized scale, 
  \item High accuracy and computational efficiency,
  \item Robustness in environments with challenging texture.
\end{itemize} 
In the current implementation, each operation of scale optimization adds only about 2 to 3 milliseconds overhead (with around 2000 points) on average when extending a monocular VO to a stereo VO. We have evaluated an extended stereo VO using the proposed method on standard public datasets, as well as our own (publicly-available) datasets. Using the standard public datasets, we demonstrate that the proposed method achieves accuracy comparable to the state-of-the-art stereo matching-based VO with much less computational cost. In scenarios with challenging textures, the performance of state-of-the-art stereo VO degrades, while the proposed method performs sufficiently well without significant degradation of accuracy or performance (see Sec.~\ref{sec:experiments}). An open-source implementation of this work is available online\footnote{\url{https://github.com/jiawei-mo/scale_optimization}}.

\section{Related Work}
\label{sec:related_work}
Stereo VO has been widely explored, with many approaches~\cite{cvivsic2015stereo,engel2015large, pire2017s, gomez2016robust, mur2017orb} relying on stereo matching. S-PTAM \cite{pire2017s} is one of the recent developments in stereo VO, which extends~PTAM \cite{klein2007parallel} to a stereo system by using stereo matching to generate new 3D points. Stereo ORB-SLAM \cite{mur2017orb} is another example of stereo VO that depends on stereo matching. Engel et al. extended their monocular LSD-SLAM~\cite{engel2014lsd} to a stereo VO~\cite{engel2015large}. Monocular LSD-SLAM is purely based on direct method (directly minimizing photometric error, independent of feature matching), but as LSD-SLAM uses stereo matching, it is no longer a fully direct method. VO algorithms with stereo matching often suffer from the problems discussed in Sec.~\ref{sec:introduction}. They tend to fail if the scene texture is repetitive, and are not computationally efficient.

The stereo matching methods mentioned above mainly focus on the patch appearance (\textit{e.g.}, normalized cross-correlation or feature descriptor) to determine stereo correspondence, referred to as `local' methods. To improve the robustness of stereo matching, authors have looked at global stereo matching for VO which exploits non-local constraints such as smoothness. One example is the stereo VO developed by Stereolab for their ZED stereo camera~\cite{zedcam}. While the localization accuracy of this approach could be improved, real-time performance is achieved by performing stereo matching on a GPU. This adds to energy consumption and increases system complexity, which is not desirable for mobile robots.

Forster et al. extended their monocular SVO~\cite{forster2014svo} for multi-camera systems~\cite{forster2017svo}, though not particularly for a stereo camera. Instead of stereo matching, they couple all
cameras into one function to reduce photometric error. This error function is calculated by projecting 3D points onto all visible image frames. The accuracy is further improved and the scale problem is solved implicitly. However, computational cost significantly increases because of this augmented error function. Stereo DSO~\cite{wang2017stereo} is a hybrid model, which uses stereo matching to initialize depth for each keyframe; the stereo image is also coupled into the error function. In spite of the computational cost, Stereo DSO is a highly accurate approach to VO.

\section{Methodology}
\label{sec:methodology}
\begin{figure}
    \centering
    \includegraphics[width=\textwidth]{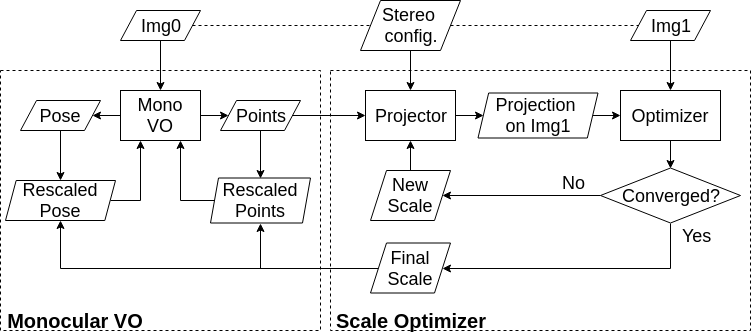}
    \caption{Method overview. The two components, namely the Monocular VO (left) and the Scale Optimizer (right), run on two different cameras of the stereo pair. The Monocular VO tracks camera pose and reconstructs 3D points, whose scale is estimated/optimized by the Scale Optimizer. }
    \label{fig:overview}
\end{figure}
Fig.~\ref{fig:overview} shows an overview of the proposed algorithm. For the current implementation, we adopt DSO~\cite{engel2018direct} to perform monocular VO and enhance it to a two-camera system using the proposed scale optimization method. However, any monocular VO algorithm can be used in this step. DSO was chosen for two reasons. First, as of the time of writing, DSO demonstrates state-of-the-art accuracy among monocular VO methods. The accuracy of the extended stereo VO using scale optimization is strongly dependent on the underlying monocular VO. Second, DSO is one of the few existing monocular VOs which are purely based on direct method. Since scale optimization is also purely based on direct method, the extended stereo VO is thus purely based on direct method.

\paragraph*{Notation}
Being consistent with DSO, we use lower-case letters ($d$) to represent scalars, bold lower-case letters ($\mathbf{t}$) to represent vectors, bold upper-case letters ($\mathbf{R}$) to represent matrices, and upper-case letters ($I$) to represent functions.

\subsection{Monocular VO}
\label{sec:mono_vo}
As shown in Fig.~\ref{fig:overview}, we use DSO as our Monocular VO to track camera poses and generate 3D points. Here we will briefly introduce DSO and then only focus on the components that are related to the scale. Readers are referred to~\cite{engel2018direct} for other details. 

DSO is based on direct method, camera poses are tracked by minimizing the photometric error. Being independent of feature description and matching gives direct visual odometry the potential of running at high frame rates and makes it robust to repetitive texture. These advantages are inherited by the extended stereo VO with scale optimization. DSO is a keyframe-based VO. Bundle adjustment of camera poses and 3D points are conducted only for keyframes~\cite{strasdat2010real}. The other frames are tracked with respect to keyframes, and they are used to refine the 3D points (inverse depth in DSO). Therefore, the proposed scale optimization is only called for keyframes. This further reduces the overhead of extending DSO to a stereo system using scale optimization.

In DSO, the following error function\footnote{The affine brightness terms in Eq. (\ref{eq:photo_error}) are ignored for simplicity.} is used at each keyframe to optimize all camera poses and 3D points within the current sliding window:
\begin{equation}
    E_{photo} = \sum_{i\in \textit{F}}\sum_{\textbf{p}\in P_i}\sum_{j\in obs(\textbf{p})}E_{\textbf{p}j} 
\label{eq:dso_error}
\end{equation}
\begin{equation}
    E_{\textbf{p}j} = \sum_{\textbf{p}\in N_{\textbf{p}}}w_{\textbf{p}} ||I_j[\Pi_0(\textbf{R}\Pi_0^{-1}(\textbf{p},d_\textbf{p})+\textbf{t})]-I_i[\textbf{p}]||_\gamma
\label{eq:photo_error}
\end{equation}

However, Eq. (\ref{eq:photo_error}) is invariant to scale. If we re-scale the translation $\mathbf{t}$ and 3D points $\Pi_0^{-1}(\textbf{p}, d_\textbf{p})$ with a factor of $s$, the Eq. (\ref{eq:photo_error}) is unchanged:
\begin{align*}
    E'_{\textbf{p}j} = \sum_{\textbf{p}\in N_{\textbf{p}}}w_{\textbf{p}} ||I_j[\Pi_0(\textbf{R}\cdot s\Pi_0^{-1}(\textbf{p}, d_\textbf{p})+s\textbf{t})]-I_i[\textbf{p}]||_\gamma \\
    = \sum_{\textbf{p}\in N_{\textbf{p}}}w_{\textbf{p}} ||I_j[\Pi_0(s(\textbf{R}\Pi_0^{-1}(\textbf{p}, d_\textbf{p})+\textbf{t}))]-I_i[\textbf{p}]||_\gamma = E_{\textbf{p}j}
\end{align*}
Thus, monocular DSO is unaware of scale. With more error coming into the system, the scale tends to drift. An example can be seen in Fig.~\ref{fig:scale_effect}. Readers are referred to~\cite{engel2015large} for notations and detailed treatments. 

Stereo DSO~\cite{wang2017stereo} solved the scale problem by using stereo matching to initialize depth and extending the error term (\ref{eq:dso_error}) to:
\begin{equation*}
    E_{photo} = \sum_{i\in \textit{F}}\sum_{\textbf{p}\in P_i}(\sum_{j\in obs(\textbf{p})}E_{\textbf{p}j} + \lambda E_{\textbf{p}s}) 
\end{equation*}
where $E_{\textbf{p}s}$ is the photometric error when projecting onto the stereo frame. It is coupled into the system by a weight of $\lambda$. The scale problem is implicitly solved by integrating the stereo baseline into the error function. Stereo DSO exhibits high accuracy, but its computational cost is much higher than that of monocular DSO. It adopts stereo rectification for stereo matching, but stereo rectification itself is computationally slow. Also, stereo matching makes Stereo DSO not fully based on direct method. 

\subsection{Scale Optimization}
With the goal of solving the scale issue of monocular VO with minimal computational cost, while still being robust in challenging, texture-depleted environments, we propose our scale optimization method that extends a monocular VO to a stereo VO effectively and efficiently. As Fig.~\ref{fig:overview} shows, scale optimization is of modular design which makes it trivial to integrate this into any existing monocular VO algorithm. The inputs to the scale optimization are the 3D points from monocular VO, and the output is the optimized scale of the current frame. The optimized scale is then integrated back into the monocular VO for scale adjustment.

For each keyframe, DSO performs bundle adjustment to optimize the camera poses and 3D points jointly. Subsequently, the optimized 3D points are handed over to the scale optimizer. They are projected onto the stereo frame (\textit{Img1} in Fig.~\ref{fig:overview}) to find an optimal scale such that the photometric error is minimized:
\begin{equation}
    E_{scale} = \sum_{\textbf{p}\in P}E_{\textbf{p}scale} 
\label{eq:scale_error}
\end{equation}
\begin{equation}
    E_{\textbf{p}scale} = w_{\textbf{p}} ||I_1[\Pi_1(\textbf{T}_{stereo}\cdot s\Pi_0^{-1}(\textbf{p},d_\textbf{p}))]-I_0[\textbf{p}]||_\gamma
\label{eq:scale_proj}
\end{equation}
For each pixel $\mathbf{p}$ with its depth $d_\mathbf{p}$, it is first back-projected to 3D space by $\Pi_0^{-1}(\textbf{p},d_\textbf{p}))$, then it is re-scaled by current scale $s$. The re-scaled 3D point is transformed to the stereo camera coordinate by $\mathbf{T}_{stereo}$, which is projected onto the stereo image frame by $\Pi_1$. The photometric error is calculated as the Huber norm of pixel intensity difference. 

The error term in Eq. (\ref{eq:scale_proj}) is simplified compared to the error term in Eq. (\ref{eq:photo_error}). Eq. (\ref{eq:scale_proj}) only focuses on the exact pixel at the projection instead of a pattern around the projection as in Eq. (\ref{eq:photo_error}), in order to further reduce computational cost. The projection $\Pi_1$  in Eq. (\ref{eq:scale_proj}) is parameterized by the relative pose between the stereo cameras ($\mathbf{T}_{stereo}$) and current scale of the 3D points ($s$) (\textit{i.e.}, \textit{Stereo config.} and \textit{New Scale} in Fig.~\ref{fig:overview}). $\mathbf{T}_{stereo}$ is pre-calibrated and fixed, so the scale $s$ is the only variable in the system. Thus, focusing on a single pixel is feasible for scale optimization, which is validated in Sec.~\ref{sec:experiments}. The necessity of using the pattern in Eq. (\ref{eq:photo_error}) is due to its high degrees of freedom including all camera poses and depths.

We use Gauss-Newton optimization~\cite{ruszczynski2006nonlinear} to solve Eq.~(\ref{eq:scale_error}). We write the photometric residual as:
\begin{align*}
    r_{\textbf{p}scale} = I_1[\Pi_1(\textbf{T}_{stereo}\cdot s\Pi_0^{-1}(\textbf{p},d_\textbf{p}))]-I_0[\textbf{p}]\\
    = I_1[\Pi_1(s\cdot \textbf{R}_{stereo}[x,y,z]^T+\textbf{t}_{stereo})]-I_0[\textbf{p}] \\
    = I_1[\Pi_1(s\cdot [x',y',z']^T+[t_x, t_y, t_z]^T)]-I_0[\textbf{p}] \\
    = I_1[
    \begin{bmatrix}
    f_x & 0 & c_x \\
    0 & f_y & c_y \\
    0 & 0 & 1 \\
    \end{bmatrix}
    \cdot
    \begin{bmatrix}
    s\cdot x' + t_x \\
    s\cdot y' + t_y \\
    s\cdot z' + t_z \\
    \end{bmatrix}
    ]-I_0[\textbf{p}] \\
    \doteq I_1[
    \begin{bmatrix}
    \frac{s\cdot f_x x' + f_x t_x}{s\cdot z' + t_z} + c_x\\
    \frac{s\cdot f_y y' + f_y t_y}{s\cdot z' + t_z} + c_y
    \end{bmatrix}
    ]-I_0[\textbf{p}]
\label{eq:scale_res}
\end{align*}
where $f_x, f_y, c_x, c_y$ are intrinsic parameters of \textit{Img1} in Fig.~\ref{fig:overview}; $[x', y', z']^T$ is the 3D point rotated by $\mathbf{R}_{stereo}$; and $\mathbf{t}_{stereo}=[t_x, t_y, t_z]^T$. The Jacobian of $r_{\mathbf{p}scale}$ with respect to the scale $s$ is:
\begin{equation*}
    \mathbf{J_s} = \frac{\partial I_1}{\partial \Pi_1} \cdot \frac{\partial \Pi_1}{\partial s}
\end{equation*}
$\frac{\partial I_1}{\partial \Pi_1}$ is the image gradient at the projection on \textit{Img1},
\begin{equation*}
    \frac{\partial \Pi_1}{\partial s} = \frac{1}{(s\cdot z' + t_z)^2}
    \begin{bmatrix}
    f_x x' t_z - f_x z' t_x\\
    f_y y' t_z - f_y z' t_y
    \end{bmatrix}
\end{equation*}

At each iteration, the Gauss-Newton algorithm will solve the above system and get a scale increment. The new scale is updated as $s_{new} = s_{old} + s_{inc}$. After convergence, the final scale is fed back into the monocular VO. Consequently, the scale of the system is accurate and consistent.

One note is that the inverse compositional method~\cite{brooks2006generalizing} is not feasible for scale optimization, because changing the scale of 3D points does not change its projection onto the original image (\textit{Img0}).

We use image pyramids to optimize scale from coarse to fine. Coarse-to-fine strategy is especially useful for the first keyframe, where the scale is totally unknown. Alternatively, stereo matching could be called at the first frame to initialize the scale.

Stereo correspondences are implicitly found all at once by optimizing the scale. Compared to explicit stereo matching, the proposed method is more robust to challenging scenes as the 3D points are already partially reconstructed by the monocular VO. Integrating them in a single error function (Eq.~(\ref{eq:scale_error})) for scale optimization is more robust than individual stereo matching, especially when the scene has repetitive textures. Experimental evaluations described in the following section further underscore this point.

\section{Experimental Evaluation}
\label{sec:experiments}
We evaluate the accuracy and efficiency of scale optimized VO through a number of experiments. These include tests on two publicly-available datasets: the KITTI Visual Odometry dataset~\cite{geiger2013vision} and the EuRoC MAV dataset~\cite{burri2016euroc}. We compare our extended DSO using scale optimization against Stereo DSO~\cite{wang2017stereo}. For naming convenience, we refer to the extended DSO using scale optimization as \textbf{SO-DSO}. For Stereo DSO, only third-party implementations are available, since the original authors are yet to publish their code, leading us to choose the best-performing\footnote{\url{https://github.com/JingeTu/StereoDSO}} among these. This particular implementation achieves reasonably high accuracy in our experiments. We choose Stereo DSO to compare with so that both algorithms use the same camera tracking algorithm (DSO), which makes it possible to directly compare stereo matching/coupling with scale optimization. We compare the accuracy of visual odometry, as well as the extra cost as stereo systems over the monocular DSO. We adopt the evaluation method used in KITTI VO benchmark, which evaluates the accuracy of the trajectories with different lengths. When testing the run-time, we use the default (slowest) setting of DSO (2000 active points, 7 max keyframes, etc.), not enforcing real-time performance, in order to maximize the accuracy. For clarity, we focus on the run-time of different components between SO-DSO and Stereo DSO, which are the scale optimizer in SO-DSO, stereo matching in Stereo DSO, and the different cost function in bundle adjustment. The experiments are carried out on a single thread of an Intel i7-6700 CPU.

\subsection{KITTI VO Dataset}
KITTI VO Dataset has $22$ driving sequences. The vehicle drives around local communities and highways capturing stereo image sequences. The ground truth is provided by a Velodyne laser scanner and a GPS localization system. However, only the first $11$ sequences are publicly available with ground truth; the ground truth of Sequences $11$ to $21$ are reserved for test and ranking of different VO algorithms. We present results from the first $11$ sequences for comparison since we do not have full access to the errors on test sequences $11$ to $21$.

\begin{table}
    \centering
    \small
    \begin{tabular}{|p{0.5cm}|p{0.9cm}|p{0.9cm}|p{0.9cm}|p{1cm}|p{1cm}|p{1.1cm}|}
        \hline
         Seq. & $t_{rel}$ \newline (\%) & $r_{rel}$ \newline (deg) & S.O. S.M. \newline (ms) & BA \newline (ms) & TPF \newline (ms) & Pts\\
        \hline
         00   & 1.35 \textbf{0.83} & 0.27 0.27 & \textbf{2.25} 12.40 & \textbf{124.07} 147.63 & \textbf{141.95} 189.88 & 2164.28 1658.76\\
        \hline
         01   & 2.72 \textbf{1.78} & 0.13 \textbf{0.11} & \textbf{1.85} 10.75 & \textbf{66.38}  73.32 & \textbf{76.42} 123.27  & 1437.18 1133.11\\
        \hline
         02   & 1.10 \textbf{0.79} & 0.22 \textbf{0.21} & \textbf{2.23} 11.28 & 121.42 \textbf{111.71} & \textbf{164.16} 171.52 & 2019.06 1426.85\\
        \hline
         03   & 3.17 \textbf{1.01} & \textbf{0.15} 0.16 & \textbf{2.44} 11.34 & 115.32 \textbf{109.05} & \textbf{97.78} 109.47 & 2241.95 1592.40\\
        \hline
         04   & 1.73 \textbf{1.01} & 0.21 \textbf{0.19} & \textbf{2.09} 11.15 &  \textbf{87.40} 104.69 & \textbf{122.44} 160.64 & 1926.45 1599.01\\
        \hline
         05   & 1.69 \textbf{0.82} & 0.20 \textbf{0.18} & \textbf{2.11} 11.20 &  \textbf{108.60} 113.93 & \textbf{119.62} 145.99 & 2028.98 1647.78\\
        \hline
         06   & \textbf{1.66} 9.19 & 0.19 \textbf{0.17} & \textbf{2.09} 11.05 &  \textbf{85.05} 90.44 & \textbf{110.03} 125.06 & 1718.27 1270.94\\
        \hline
         07   & 2.50 \textbf{1.03} & \textbf{0.32} 0.33 & \textbf{2.22} 10.66 & \textbf{113.50} 119.33 & \textbf{113.77} 134.84 & 2153.60 1716.57\\
        \hline
         08   & 1.72 \textbf{1.04} & \textbf{0.26} 0.27 & \textbf{2.08} 11.15 &  \textbf{109.24} 118.36 & \textbf{126.58} 155.55 & 1945.50 1497.82\\
        \hline
         09   & 1.88 \textbf{0.98} & 0.22 \textbf{0.19 }& \textbf{2.04} 11.22 &  \textbf{98.95} 102.23 & \textbf{130.83} 150.79 & 1900.30 1457.92\\
        \hline
         10   & 1.02 \textbf{0.61} & 0.21 \textbf{0.19} & \textbf{1.89} 10.60  &  \textbf{88.84}  93.38 & \textbf{102.70} 117.87 & 1775.74 1357.89\\
        \hline
    \end{tabular}
    \caption{Error and run-time comparisons on the KITTI dataset. For each sequence, \textbf{the upper line is the result of SO-DSO, and the lower line is for Stereo DSO}. $t_{rel}$ is translational RMSE(\%); $r_{rel}$ is rotational RMSE (degree per 100m). Results are averaged over $100m$ to $800m$ intervals. S.O. is the run-time of scale optimization; S.M. is the run-time of stereo matching; BA is the bundle adjustment run-time; TPF is the time per frame (not just keyframe); Pts is the number of 3D points in the bundle adjustment.}
    \label{tab:kitti_error}
\end{table}

Table~\ref{tab:kitti_error} shows the comparison between SO-DSO and Stereo DSO. It is worth noting that the errors of the third party implementation are quite close to the errors reported in the original Stereo DSO paper~\cite{wang2017stereo} (The error of Seq. 06 corresponds to the Figure. 4 in~\cite{wang2017stereo} with coupling factor 1). In most cases, Stereo DSO achieves higher translational accuracy. This is expected because SO-DSO depends on monocular DSO for generating 3D points, but monocular DSO is not designed for quick camera movement or low camera frame-rate (10Hz). Even worse, there are lots of sharp turns in many sequences. In Stereo DSO, integrating static stereo drastically increases accuracy and robustness for these challenging cases. We suggest static stereo for those cases where monocular VO does not work well. Nevertheless, the accuracy of SO-DSO is comparable to that of Stereo DSO. 

On the efficiency factor, however, scale optimization is approximately $5$ times faster than stereo matching. In the current implementation, we use $4$ image pyramids for robustness. After the scale is initialized, it is not necessary to use as many pyramids; thus, run-time can be further reduced. On the other hand, Stereo DSO maintains a lower number of 3D points. Stereo matching does not work well for repetitive textures, neither can it triangulate points far away since there is no disparity. The KITTI dataset has plants and far-away objects, which could be challenging for stereo matching. With more points to optimize, the bundle adjustment in SO-DSO is still faster than the one in Stereo DSO. One reason is that Stereo DSO projects points onto both stereo frames, the number of error terms is drastically increased (while improving accuracy). As a system, SO-DSO spends less time per frame (TPF), even with more points, than Stereo DSO. One point to note is that the majority of TPF is taken by monocular DSO. In theory, the overhead of SO-DSO over monocular DSO is the time taken for scale optimization (plus the negligible time needed for accessing 3D points and scale adjustment). Additionally, Stereo DSO requires data pre-processing of stereo rectification, which is also time-consuming.

\begin{figure}
    \centering
    \includegraphics[width=\textwidth]{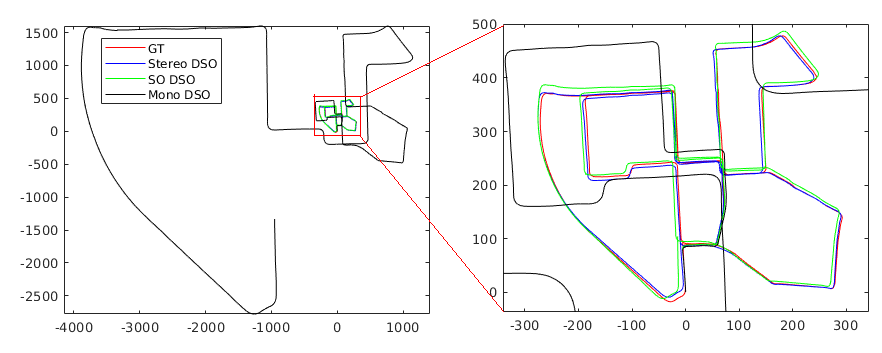}
    \caption{Effect of scale optimization on KITTI Seq. 00. Trajectories of ground truth (GT), Stereo DSO, SO-DSO, and monocular DSO are shown.}
    \label{fig:scale_effect}
\end{figure}

Fig.~\ref{fig:scale_effect} demonstrates the effectiveness of scale optimization qualitatively. We initialize the scale of monocular DSO at the beginning only with no further scale optimization, which is labeled as Mono DSO. The trajectory is close to ground truth at the beginning, but completely deviates as the scale drift becomes extremely large. However, with scale optimization throughout, the trajectory (SO-DSO) is always close to the ground truth, though not as close as Stereo DSO.

\subsection{EuRoC MAV Dataset}
We also compare SO-DSO with Stereo DSO on the EuRoC dataset. The dataset was recorded by a drone in two scenarios, Machine Hall and Vicon Room. The ground truth for Machine Hall was measured by a Leica MS50 laser tracker, which contains 3D position only. Thus, no rotational error is measured in Machine Hall tests. The ground truth for the Vicon Room was measured by a Vicon motion capture system obtaining both position and orientation.

\begin{table}[t]
    \centering
    \small
    \begin{tabular}{|p{1.1cm}|p{0.9cm}|p{0.9cm}|p{0.9cm}|p{1.1cm}|p{0.9cm}|p{1.1cm}|}
        \hline
         Seq. & $t_{rel}$ \newline (\%) & $r_{rel}$ \newline (deg) & S.O. S.M. \newline (ms) & BA \newline (ms) & TPF \newline (ms) & Pts \\
        \hline
         MH\_01\newline easy   & \textbf{0.23} 1.51 & N/A \newline N/A & \textbf{3.45} 10.72 &  \textbf{104.69} 157.57 & \textbf{36.81} 60.40 & 2770.68 2728.09 \\
        \hline
         MH\_02\newline easy   & \textbf{0.28} 1.28 & N/A \newline N/A & \textbf{3.42} 10.74 &  \textbf{105.42} 155.26 & \textbf{35.03} 55.65 & 2734.76 2691.50 \\
        \hline
         MH\_03\newline medium   & \textbf{0.59} 1.62 & N/A \newline N/A & \textbf{3.32} 10.57 & \textbf{121.23} 178.55 & \textbf{58.24} 91.03 & 2705.65 2659.79 \\
        \hline
         MH\_04\newline hard   & \textbf{0.76} x & N/A \newline x & \textbf{3.42}  x & \textbf{125.00} x & \textbf{51.09} x & 2705.13 x \\
        \hline
         MH\_05\newline hard   & \textbf{0.63} 1.22 & N/A \newline N/A & \textbf{3.23} 10.44 & \textbf{116.70} 172.04 & \textbf{46.24} 75.05 & 2674.09 2592.49 \\
        \hline
         V1\_01\newline easy    & \textbf{1.70} 2.42 & \textbf{21.63}  22.50 & \textbf{3.53} 10.70 & \textbf{127.66} 186.72 & \textbf{52.20} 84.04 & 2715.68 2654.30 \\
        \hline
         V1\_02\newline medium    & \textbf{0.78} 2.66 &  \textbf{7.06}  43.98 & \textbf{3.08} 10.27 & \textbf{133.08} 208.00 & \textbf{92.49} 139.12 & 2720.20 2709.03 \\
        \hline
         V1\_03\newline hard    & x \newline x & x \newline x & x \newline x & x \newline x & x \newline x & x \newline x \\
        \hline
         V2\_01\newline easy    & \textbf{0.57} 13.85&  \textbf{7.14} 137.62 & \textbf{3.51} 9.87 & \textbf{136.74} 158.41 & \textbf{47.21} 63.85 & 2756.59 2232.31 \\
        \hline 
         V2\_02\newline medium    & \textbf{2.50} 9.63 &  \textbf{6.76} 152.00 & \textbf{3.61} 9.77 & \textbf{132.38} 185.18 & \textbf{84.83} 102.70 & 2780.82 2487.47 \\
        \hline
         V2\_03\newline hard   & x \newline x & x \newline x & x \newline x & x \newline x & x \newline x & x \newline x \\
        \hline
    \end{tabular}
    \caption{Error and run-time comparison on EuRoC. Same notation as in Table~\ref{tab:kitti_error}, except that results are averaged over $10m$ to $80m$ intervals.}
    \label{tab:euroc_error}
\end{table}

The results are given in Table~\ref{tab:euroc_error}. For Machine Hall tests, both Stereo DSO and SO-DSO work well, at least for `easy' and `medium' tests. One example of SO-DSO on Machine Hall tests is given in Fig.~\ref{fig:mh01}. Compared with KITTI tests, the error of SO-DSO, in this case, is lower. One factor is the high frame rate (20Hz) of EuRoC dataset. On the other hand, averaging results over smaller intervals could be the reason that Stereo DSO has a higher error rate. For Vicon Room tests, neither work for `hard' tests; SO-DSO works for `easy' and `medium' tests with low error, while Stereo DSO has a noticeable reduction in accuracy. The possible reasons are image blur due to fast camera motion and poor illumination. Note that Stereo DSO has relatively more 3D points in bundle adjustment for EuRoC than that it has in KITTI. The scale of Machine Hall/Vicon Room is smaller than the street scenes in KITTI, which means more points are within the stereo camera range. From Table~\ref{tab:euroc_error}, we see that the accuracy of SO-DSO is comparable to, if not better than, Stereo DSO (with the 3rd party implementation).

\begin{figure*}[ht]
\begin{subfigure}[t]{0.32\textwidth}
    \centering
    \includegraphics[width=\textwidth]{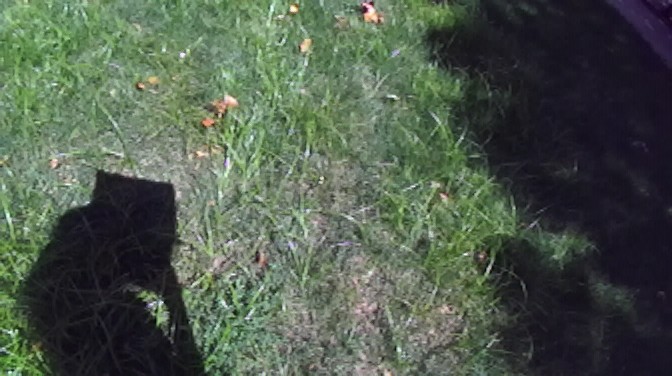}
    \caption{A view of the grass dataset.}
    \label{fig:zed_view}
\end{subfigure}
~
\begin{subfigure}[t]{0.32\textwidth}
    \centering
    \includegraphics[width=\textwidth]{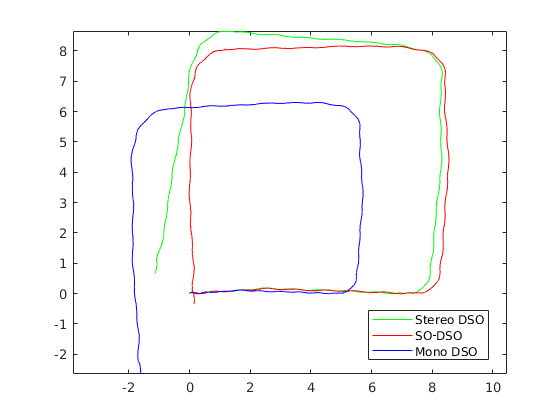}
    \caption{Trajectories (top view, in meters) estimated by the three algorithms. }
    \label{fig:zed_traj}
\end{subfigure}
~
\begin{subfigure}[t]{0.32\textwidth}
    \centering
    \includegraphics[width=\textwidth]{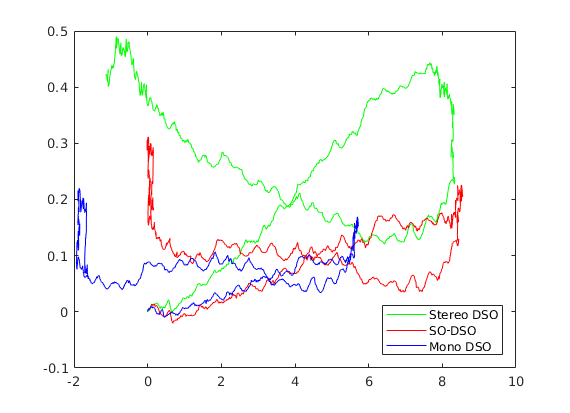}
    \caption{Elevation changes in the trajectories (in meters) estimated by the three algorithms.}
    \label{fig:zed_height}
\end{subfigure}
\caption{ZED camera experiment, ground-truth is approximately a zero-elevation, $8m \times 8m$ square.}
\end{figure*}

\begin{figure*}[bht]
\begin{subfigure}[t]{0.23\textwidth}
    \centering
    \includegraphics[width=\textwidth]{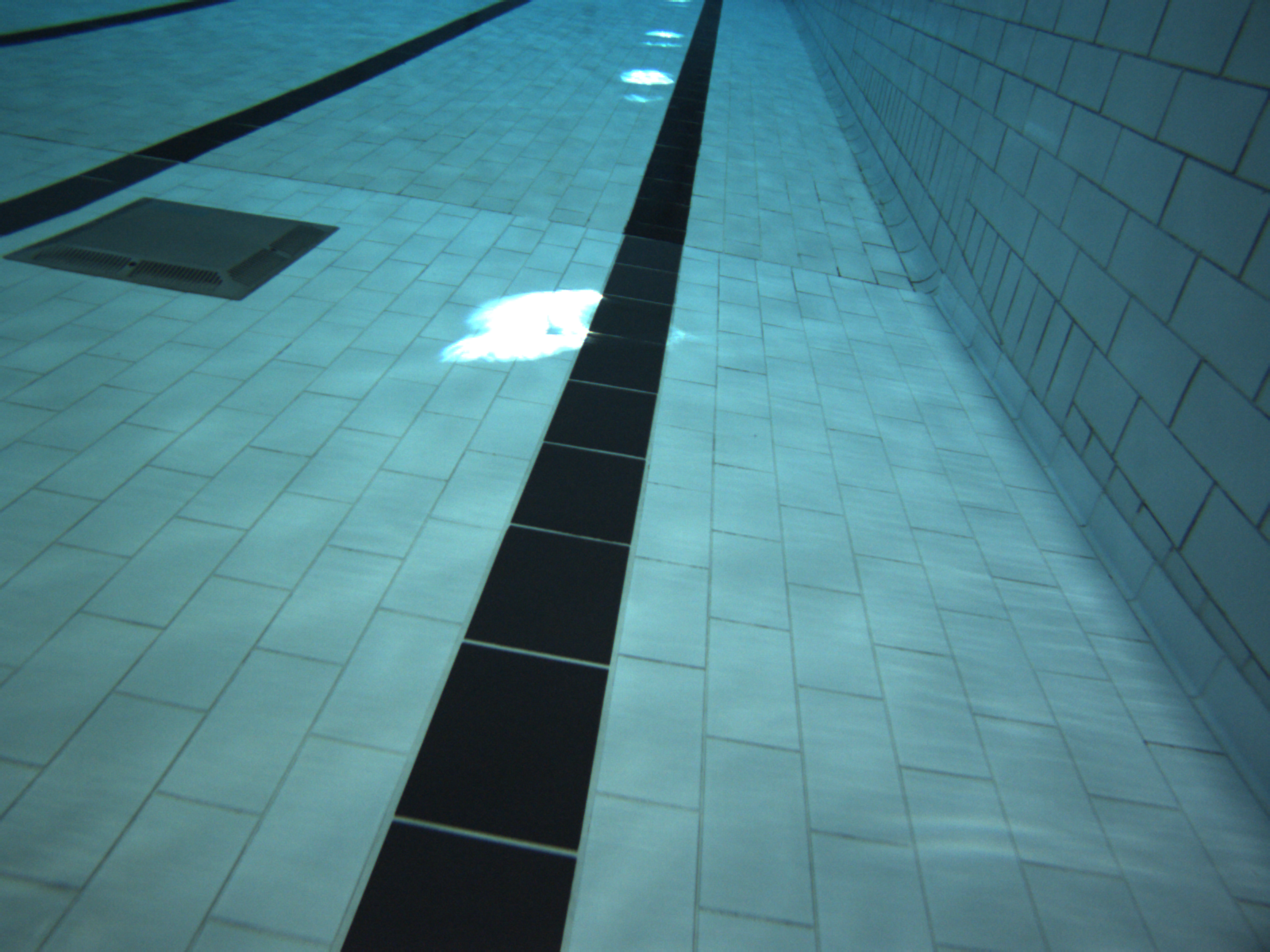}
    \caption{A snapshot of the swimming pool dataset.}
    \label{fig:pool_view}
\end{subfigure}
~
\begin{subfigure}[t]{0.23\textwidth}
\centering
\includegraphics[width=\textwidth]{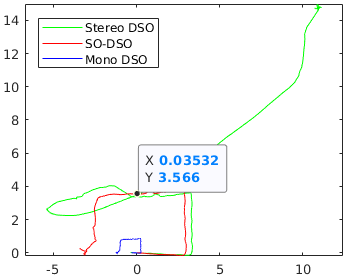}
\caption{Robot trajectories (in meters) estimated by the three algorithms in the pool. }
\label{fig:pool_top}
\end{subfigure}
~
\begin{subfigure}[t]{0.23\textwidth}
    \centering
    \includegraphics[width=\textwidth]{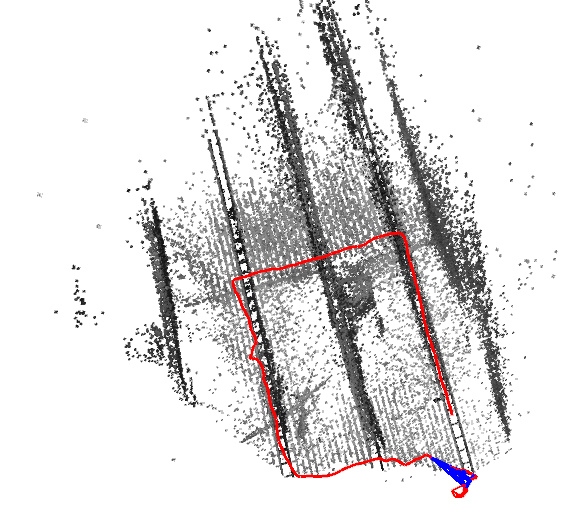}
    \caption{Reconstructed pool (top view) by SO-DSO, with robot trajectory overlaid. SO-DSO converges throughout}
    \label{fig:pool_so}
\end{subfigure}
~
\begin{subfigure}[t]{0.23\textwidth}
    \centering
    \includegraphics[width=\textwidth]{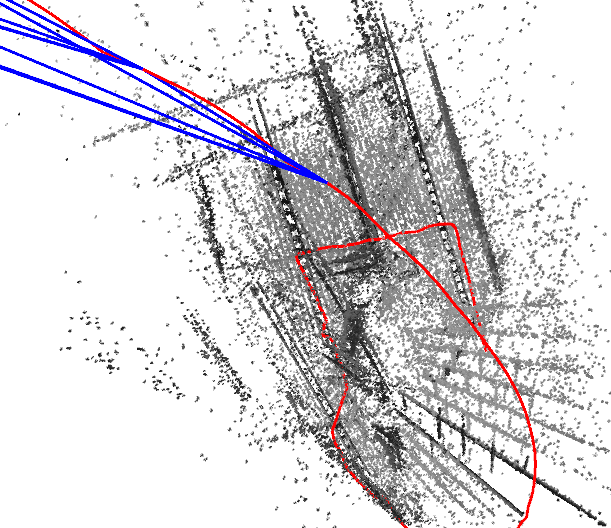}
    \caption{Reconstructed pool (top view) by Stereo DSO, with robot trajectory overlaid. Stereo DSO diverges halfway through the trajectory.}
    \label{fig:pool_sm}
\end{subfigure}
\caption{Evaluating VO in a pool environment on an AUV. The width of two swimming lanes combined is about 3.6m.}
\label{fig:sovo_pool}
\end{figure*}

Also, the high computational efficiency of our approach is further validated on this dataset. Scale optimization is faster than stereo matching, and bundle adjustment of SO-DSO is faster than that of Stereo DSO. 

The TPF drops significantly for both algorithms. The drone used in EuRoC moved slower (\textless 1m/s) than the cars in KITTI (\textgreater 4m/s), and the EuRoC dataset has a higher frame rate. The distance of camera movement is one important factor for DSO to create new keyframes. Since the camera movement is subtle, fewer keyframes are selected in EuRoC dataset. Thus, both algorithms run faster on the EuRoC dataset, with SO-DSO being the faster of the two.

After the comparison of Stereo DSO and SO-DSO on both KITTI and EuRoC datasets, it is evident that extending monocular VO using scale optimization significantly reduces computational cost without a significant loss of accuracy.

\subsection{Terrestrial Data}
\label{sec:exp_zed}
To validate the robustness of VO with scale optimization in outdoor settings, we use a ZED camera to record a stereo dataset, a snapshot of which is shown in Fig.~\ref{fig:zed_view}. The camera is carried by hand on an approximately $8m \times 8m$ square path without much elevation change. The camera is pointed at the grass throughout the trajectory. Because of the sun, the brightness changes drastically when moving into the shadows. This dataset is used intentionally to challenge stereo matching, where the grass is repetitive and of high frequency.

Fig.~\ref{fig:zed_traj} shows the results. We first run monocular DSO on this dataset, and as the blue trajectory shows, its scale is incorrect and inconsistent. If we run scale optimization on top of monocular DSO (\textit{i.e.}, SO-DSO), the trajectory is roughly a flat $8m \times 8m$ square, and returns to the start position, without significant elevation change (Figure~\ref{fig:zed_height}). On the other hand, the trajectory generated by Stereo DSO has a quite accurate scale, but the camera does not return to the start position due to rotational error. A possible reason is that the wrong stereo correspondences degrade the accuracy, which is already mentioned in the Stereo DSO paper~\cite{wang2017stereo}.

\subsection{Underwater Data}
\label{sec:exp_pool}
We also evaluate SO-DSO and Stereo DSO on a dataset we collected using the Aqua underwater robot~\cite{Sattar_ComputerMagazine} in a swimming pool, illustrated in Fig.~\ref{fig:pool_view}. This represents a challenging environment for VO methods because of the reflections that occur on the water and the lack of distinguishable visual features.

The results are shown in Fig.~\ref{fig:pool_top}. Monocular VO and SO-DSO generate two similar trajectories but with different scales. With scale optimization, the distance between the two red horizontal lines in Fig.~\ref{fig:pool_top} is around $3.566$ meters, which is close to the ground truth (which was measured to be approximately $3.6m$). Stereo DSO works quite well at the early stage but fails to converge to the ground truth towards the end, and this behavior was replicated across multiple evaluations on this underwater dataset. Fig.~\ref{fig:pool_so} and Fig.~\ref{fig:pool_sm} show the top-view of the reconstructed swimming pool environment as a qualitative comparison between the two methods. Additionally, a video demonstration of the proposed method accompanies the paper, and the datasets of Sec.~\ref{sec:exp_zed} and Sec.~\ref{sec:exp_pool} are available online\footnote{\url{https://drive.google.com/open?id=1r-vsnkythfqqq0Ly0QZ34_W9Ay8cIJPN}}.

\section{Conclusions}
In this paper, we proposed a new algorithm for extending monocular visual odometry to a stereo system. It combines the advantages of monocular visual odometry and stereo visual odometry, namely computational efficiency and scale awareness. For demonstration, the monocular DSO is used to track the camera poses and generate 3D points; while the other camera in the stereo setup is used to optimize the scale of DSO. In experimental validations on public datasets and real-world recorded data, we show the proposed scale optimization approach to be very fast and reasonably accurate, and also robust in scenes of challenging texture. 

Future extensions to this work will include monocular VO failure detection so that scale optimization and stereo matching can alter in different scenarios, in order to balance accuracy and efficiency. We also intend to include loop-closing into the extended visual odometry to further improve accuracy. 

\section*{Acknowledgment}
We gratefully acknowledge the support of the MnDRIVE initiative for this research.

\bibliographystyle{ieeetr}
\bibliography{citation}
\end{document}